# Using Complex Wavelet Transform and Bilateral Filtering for Image Denoising

Seyede Mahya Hazavei
Hamedan University of Technology
Hamedan, Iran
m.hazavei@yahoo.com

Hamid Reza Shahdoosti
Hamedan University of Technology
Hamedan, Iran
h.doosti@hut.ac.ir

*Abstract*— **The bilateral filter is a useful nonlinear filter which without smoothing edges, it does spatial averaging. In the literature, the effectiveness of this method for image denoising is shown. In this paper, an extension of this method is proposed which is based on complex wavelet transform. In fact, the bilateral filtering is applied to the low-frequency (approximation) subbands of the decomposed image using complex wavelet transform, while the thresholding approach is applied to the high frequency subbands. Using the bilateral filter in the complex wavelet domain forms a new image denoising framework. Experimental results for real data are provided, by which one can see the effectiveness of the proposed method in eliminating noise.**

## I. INTRODUCTION

In a digital image, different sources of noise exist. Some noise components exhibits non-uniform spatial properties, some of which are the photo response non-uniformity (PRNU) the dark signal non-uniformity (DSNU). This kind of noise is usually called fixed pattern noise (FPN) due to the fact that the spatial pattern of this noise does not vary with time. On the other hand, temporal noise, has a spatially varying pattern. Several examples of temporal noise are reset noise, read noise and photon shot noise [1]. Generally, many factors can affect the overall noise properties namely, pixel dimensions, sensor type, ISO speed, exposure time, and temperature.

Many denoising methods have been proposed during the years. One of the most popular approaches is wavelet thresholding. In this method, the noisy image is decomposed into its low-frequency (approximation) and high-frequency (detail) subbands firstly. Owing to the fact that, the image information is transformed in a few large coefficients, the high frequency subbands are modified using soft or hard thresholding operations [2]. The threshold value selection is the critical step in the wavelet thresholding. For this purpose, different threshold selection approaches have been introduced, such as, SureShrink [3], BayesShrink [4] and VisuShrink [5]. The VisuShrink approach uses the minimax error measure to develop a universal threshold which is a function of the noise variance and the size of image. In the SureShrink technique, the threshold value is obtained by optimizing the Stein's unbiased risk estimator. The BayesShrink method models the distribution of the wavelet coefficients as Gaussian in a Bayesian framework to determine the threshold value. By taking into account the interscale and intrascale correlations of the wavelet coefficients, these approaches have later been modified [6]–[10]. The method proposed in [7] models the adjacent of coefficients at neighborhood scales and positions as Gaussian scale mixture and exploits the Bayesian least squares estimation approach to modify the coefficients of the wavelet. This method is called BLS-GSM, and is one of the successful denoising methods usually used as a benchmark in the literature on account of its excellent performance. However, several denoising methods have surpassed the performance of BLS-GSM. For example the method proposed in [11] models the subbands of coefficients of wavelet as a multiplication of two independent uniform Gaussian Markov random fields by using a global field of Gaussian scale mixtures, and proposes an iterative denoising method. In addition, the methods proposed in [12]–[13] introduce algorithms for denoising gray-scale and color image based on redundant and sparse representations over learned dictionaries, in which training procedure exploits the K-SVD algorithm. Ref. [14] groups image patches into three dimensional arrays, and utilizes a collaborative filtering procedure including a three dimensional transformation, shrinkage of the transform coefficients, and inverse three dimensional transformation. Method proposed in [15] uses a linear combination of noisy image patches to model an ideal image patch, and forms a total least squares estimation method.

The bilateral filter is also popular denoising method [16]. Although the name "bilateral filter" was given to this method in [16], several versions of this method have been already proposed, namely, the neighborhood filter, SUSAN filter and the sigma filter. This filter calculates a weighted average of the pixels in a local neighborhood, in which the assigned weights depend on both the intensity distance and spatial distance. In this technique, details such as edges and textures are preserved while noise is suppressed.

An extension of the bilateral filter has been proposed in [1] which incorporates the bilateral filter into a wavelet framework. It was shown that the image denoising performance of the this method is better than the conventional bilateral filter. In this paper, we improve the method proposed

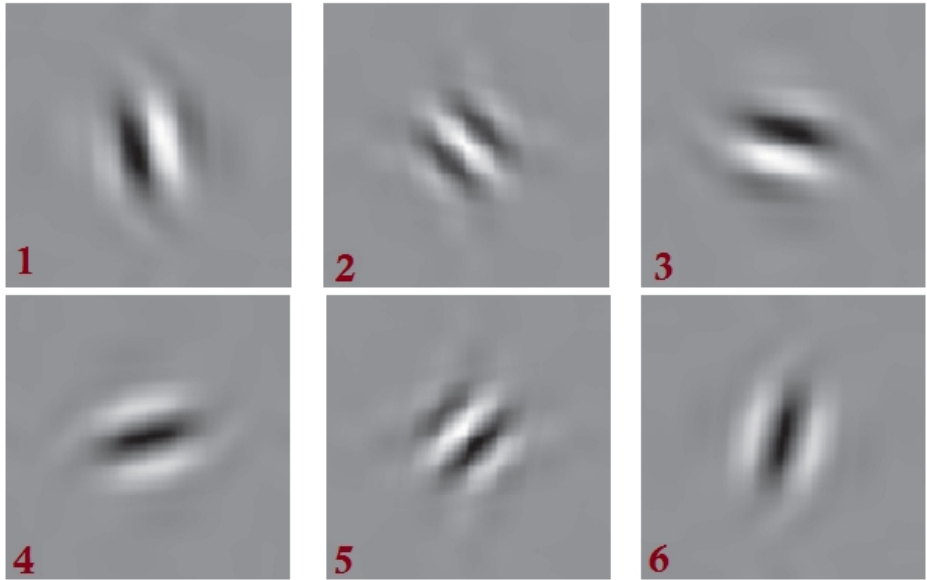

Fig. 1. Typical wavelets associated with the oriented two dimensional complex wavelet transform.

in [1], by using complex wavelet transform instead of the conventional wavelet.

The conventional wavelet suffer from shift dependence. This causes that the decomposition of signal energy between the scales of a multiresolution decomposition can vary considerably if the initial signal is shifted prior to decomposition [17]. To deal with the problem of shift variance, this types of wavelets have been introduced. A complex wavelet is composed of two real wavelets with a 90° phase difference. Hence, the complex wavelet transform, is better than the conventional wavelet due to the fact that it has good directional selectivity in two dimensions, and is approximately shift invariance. Fig. 1 shows the six directional components of the complex wavelet transform.

In Section II, we review the bilateral filter. Section III proposes a new image denoising algorithm by incorporating the bilateral filter into a complex wavelet framework. Section IV shows the capability of the proposed method by using simulations and real data experiments. Section V is devoted to the conclusion remarks.

## II. BILATERAL FILTER

Removing the noise and simultaneously retaining the important image features e.g., textures, details and edges as much as possible are the goal of image denoising. Linear filters such as Gaussian functions consisting of convolving the noise signal with a constant coefficients to obtain a linear combination of neighboring samples, have been widely exploited for noise reduction specifically, when the noisy image contains additive white Gaussian noise. However this strategy can blur and smooth the edges of the image with incomplete noise suppression and poor feature localization. Owing to the fact that not only the forward but also the inverse Fourier transforms of a Gaussian filter are real Gaussian functions, as well as their shapes are easily specified, Gaussian filters have attracted a great deal of interest. If the filter in the frequency domain becomes narrower, its response in the spatial domain will be wider which results in increased blurring/smoothing. In these types of filters, weight of the pixels decreases by distance from the center of the filter as shown by the following formula:

$$G(x,y) = \frac{1}{2\pi\sigma^2} e^{-\frac{(x^2+y^2)}{2\sigma^2}} \quad (1)$$

The main assumption of these filters is that natural images have smooth variations and adjacent pixels have similar values. So, these filters reduce the noise and simultaneously preserve the image features by averaging the adjacent pixel values over a neighborhood. However, this assumption is not realistic in the location of edges where the spatial variations are significant and the use of Gaussian filter can blur the edges. By filtering the image in both range and domain (space), the bilateral filter overcomes this problem. Bilateral filtering is a nonlinear and local method which takes into account not only the gray level (color) similarities but also the geometric distance of the neighboring pixels. The process used by the bilateral filter at a pixel location $i$ is as follows [18]:

$$BI(i) = \frac{1}{w} \sum_{j \in s} G_{\sigma_s}(\|i-j\|) G_{\sigma_r}(|I(i)-I(j)|) I(j) \quad (2)$$

where $s$ is a local window around the $i^{th}$ pixel, I is the noise image, I($i$) denotes the value of the $i^{th}$ pixel of the noisy image, $G_{\sigma_s}$ and $G_{\sigma_r}$ are Gaussian filters with standard deviations $\sigma_s$ and $\sigma_r$, respectively, and $w$ is a normalization constant [18].

The behavior of the bilateral filter is controlled by two parameters $\sigma_r$ and $\sigma_s$. The dependency between parameters $\sigma_r$ and $\sigma_s$ on the behaviors of the bilateral filter is investigated in Ref. [19]. In addition, Ref. [1] shows that the optimal parameter $\sigma_s$ is relatively insensitive to noise power in comparison with the optimal parameter $\sigma_r$ and is selected depending on the desired low-pass filtering. If $\sigma_s$ becomes larger, the edges are blurred more and more, because it considers the values of more distant pixels. Furthermore, $\sigma_s$ depends on the image size such that if the image size is scaled down or up, $\sigma_s$ must be altered in order to obtain the similar results. Empirically a good value for the parameter $\sigma_s$ is roughly [1.5–2.1], while a good value for the parameter $\sigma_r$ depends on the noise standard deviation $\sigma_n$.

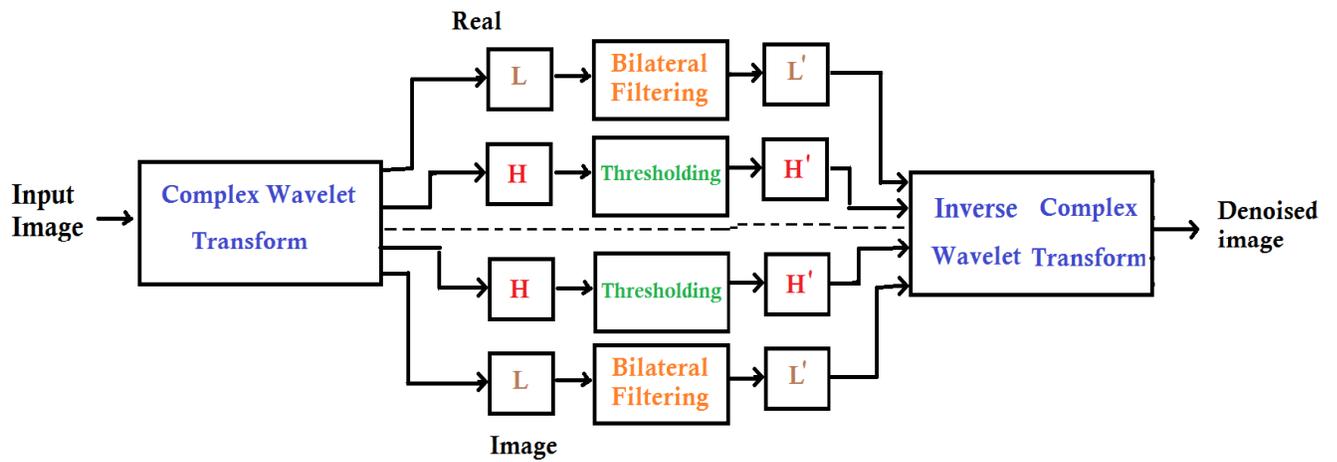

Fig. 2. The proposed image denoising framework.

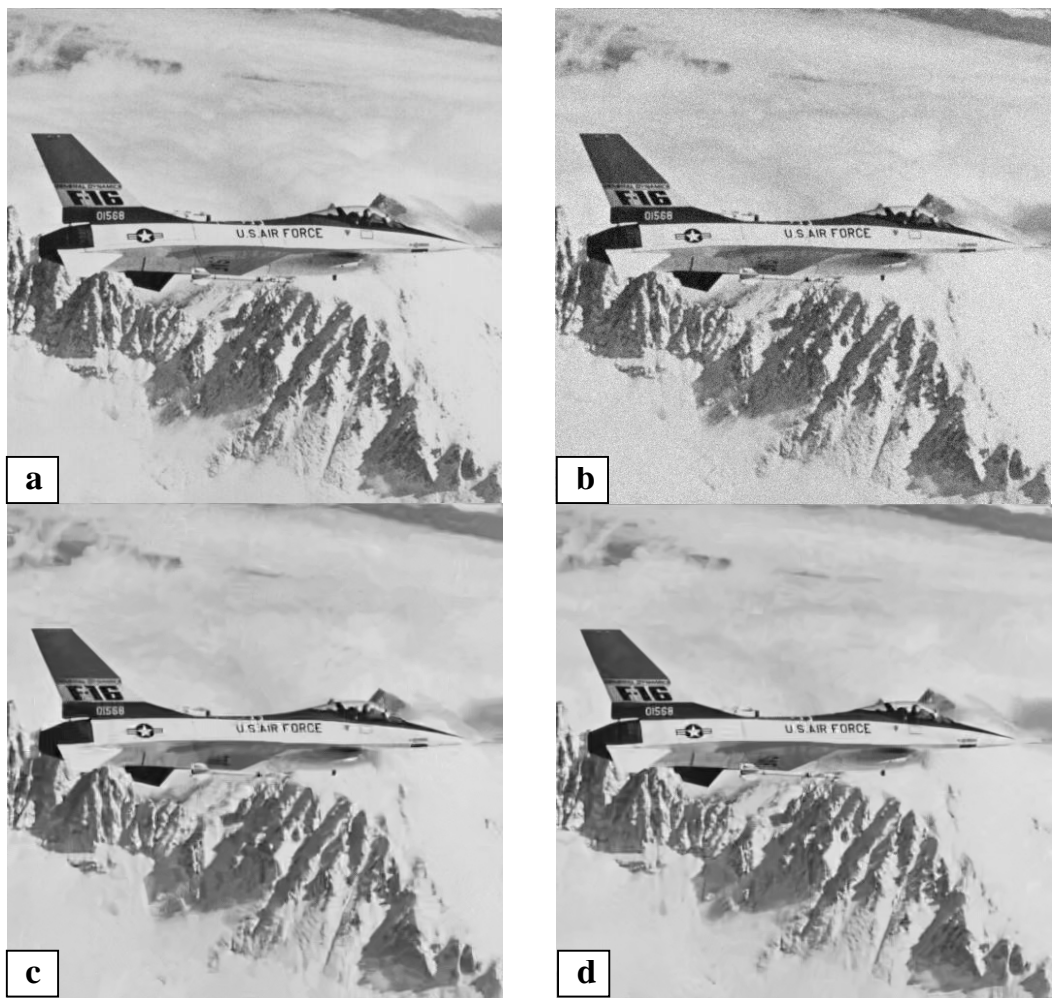

Fig. 3. (a) The Jetplane image. (b) The noisy Jetplane image. (c) The result of the method proposed in [1]. (d) The result of the proposed method.

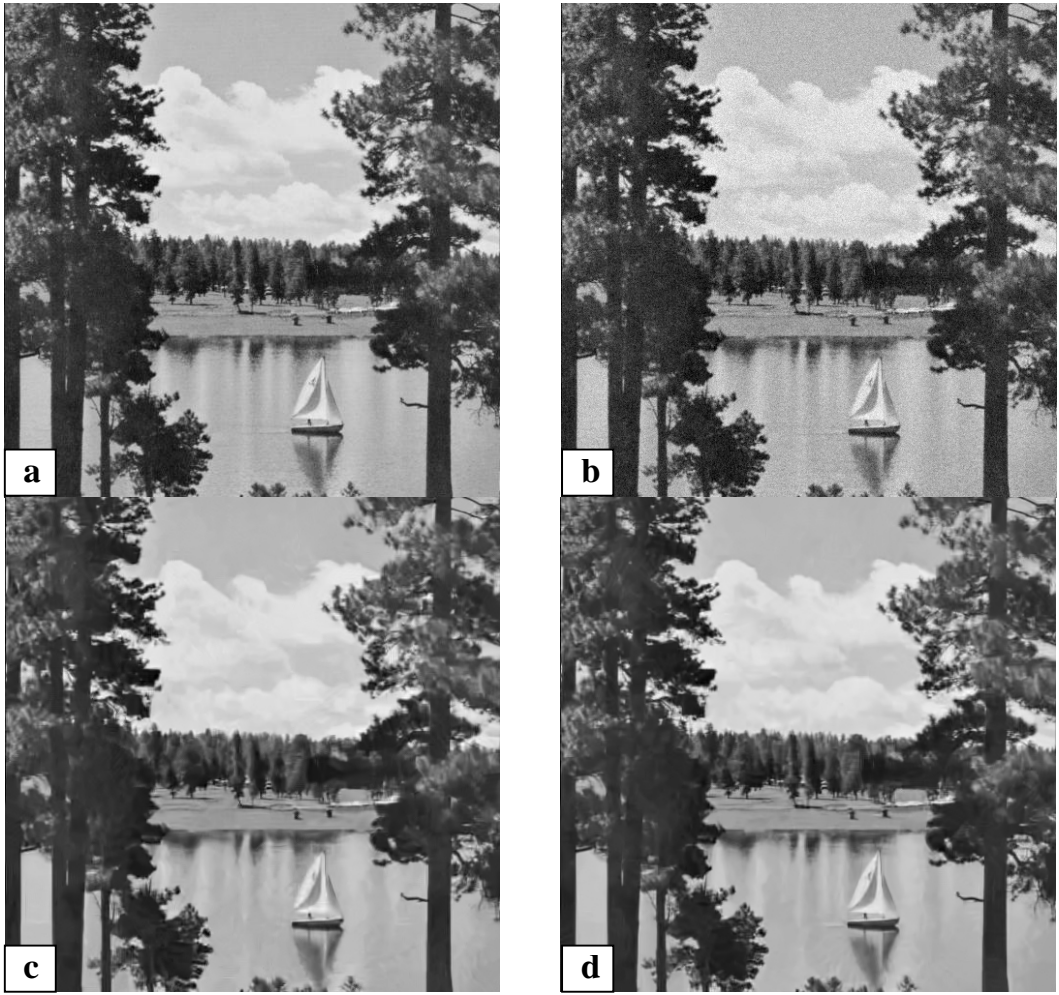

Fig. 4. (a) The Lake image. (b) The noisy Lake image. (c) The result of the method proposed in [1]. (d) The result of the proposed method.

### III. PROPOSE METHOD

For analyzing the information of images, multiresolution analysis has been proven to be an effective tool [20-23]. The noise and image information can be better distinguished at one resolution level than another. Ref. [1] showed that bilateral filter is very useful for removing the noise of low frequency subband, while thresholding is a better choice for eliminating high frequency subbands. Using the same strategy, we have proposed a new image denoising method illustrated in Fig. 2. The signal is decomposed into its different frequency subbands using complex wavelet transform; before the signal is reconstructed back, the bilateral filtering method is used to remove the noise of the approximation subbands while the thresholding method is used to denoise the high frequency subbands. In fact, the proposed image denoising framework combines bilateral filtering and thresholding methods to take advantage of both methods. In the next section, we demonstrate that the proposed framework using complex wavelet transform produces results better than the one using the conventional wavelet. This result is expected due to the advantages of the complex wavelet transform compared with conventional wavelet which are having a better directional selectivity and having the shift invariant property. The steps of the proposed method are as follows:

1- Use complex wavelet transform to decompose the noisy image into low frequency and high frequency subbands.
2- Apply the bilateral filter to the low frequency subbands. This step gives the denoised approximations.

3- Apply the wavelet thresholding to remove the noise of high frequency subbands. This step gives the denoised subbands including edges and textures.
4- Apply the inverse complex wavelet transform to the denoised subbands (high frequency and low frequency) to obtain the denoised image.

## IV. Experimental Results

In order to investigate the performance of the proposed framework both visually and quantitatively, several experiments on some famous images are conducted. This section do quantitative comparison by simulating noisy images. In the simulation, white Gaussian noise with various standard deviations is added to the standard test images. Then denoising algorithms are used to denoise images and the PSNR results are calculated. Here, five noisy images are simulated by adding white Gaussian noise with standard deviation $\sigma_n$= 10, 20, 40, 50 and 60. PSNR results for the two methods i.e., the method proposed in [1] and the proposed method are included in Table I. As can be seen from this table, the proposed method has a better performance in terms of PSNR.

For the visual comparison, two data, namely Jetplane and Lake are used which have been shown in Figs. 3 and 4, respectively. Figs. 3(a) and 4(a) show the original images. Figs. 3(b) and 4(b) shows the simulated noisy images. The denoised images for the two methods are shown in Figs. 3(c) and 3(d), and 4(c) and 4(d). As can be seen from these figures, the proposed method has a better visual performance such that the edges are better recovered and the noise is better attenuated by the proposed method.

## V. Conclusion

In this paper, we presented a new multiresolution image denoising framework, taking advantages of bilateral filtering and complex wavelet thresholding. In the proposed framework, the noisy image is decomposed into low- and high-frequency components using complex wavelet transform, and the bilateral filtering is applied on the low frequency subbands and wavelet thresholding is applied on the high frequency subbands. Then, the inverse complex wavelet transform is applied to the denoised subbands. Experiments on real images showed the effectiveness of the proposed framework.

TABLE I. PSNR for Four Widely Used Images.

| Images | Method proposed in [1] | Proposed method |
|---|---|---|
| **Barbara** | | |
| 10 | 31.79 | 32.61 |
| 20 | 27.74 | 28.55 |
| 30 | 25.61 | 26.83 |
| 40 | 23.10 | 23.96 |
| 50 | 22.56 | 23.29 |
| **Boats** | | |
| 10 | 32.58 | 33.21 |
| 20 | 29.25 | 29.87 |
| 30 | 27.24 | 28.76 |
| 40 | 25.76 | 26.30 |
| 50 | 24.63 | 25.05 |
| **Lake** | | |
| 10 | 31.33 | 32.14 |
| 20 | 28.40 | 29.06 |
| 30 | 26.57 | 26.93 |
| 40 | 24.21 | 24.84 |
| 50 | 23.36 | 23.81 |
| **Jetplane** | | |
| 10 | 33.27 | 33.88 |
| 20 | 30.18 | 31.06 |
| 30 | 28.20 | 28.79 |
| 40 | 25.95 | 26.41 |
| 50 | 24.69 | 25.16 |